\documentclass[letterpaper]{article} 
\usepackage{aaai25}  
\usepackage{times}  
\usepackage{helvet}  
\usepackage{courier}  
\usepackage[hyphens]{url}  
\usepackage{graphicx} 
\usepackage{natbib}  
\usepackage{caption} 
\usepackage{algorithm}
\usepackage{algorithmic}

\usepackage{algorithm}
\usepackage{algorithmic}
\usepackage{amsmath}
\usepackage{multirow}
\usepackage{comment}
\usepackage{graphicx}
\usepackage{makecell}
\usepackage{amssymb}
\usepackage{dsfont} 
\usepackage{booktabs}
\usepackage{xcolor}

\usepackage{newfloat}
\usepackage{listings}
\DeclareCaptionStyle{ruled}{labelfont=normalfont,labelsep=colon,strut=off} 
\floatstyle{ruled}
\newfloat{listing}{tb}{lst}{}
\floatname{listing}{Listing}
\title{BGDB: Bernoulli-Gaussian Decision Block with Improved Denoising Diffusion Probabilistic Models}
\author{
    Chengkun Sun\textsuperscript{\rm 1},
    Jinqian Pan\textsuperscript{\rm 1},
    Russell Stevens Terry\textsuperscript{\rm 2},
    Jiang Bian\textsuperscript{\rm 1},
    Jie Xu\textsuperscript{\rm 1}
}
\affiliations {
    \textsuperscript{\rm 1}Department of Health Outcomes and Biomedical Informatics, University of Florida, Gainesville, FL 32611, USA\\
    \textsuperscript{\rm 2}Department of Urology, University of Florida, Gainesville, FL 32611, USA\\
    sun.chengkun@ufl.edu,
    jinqianpan@ufl.edu,
    russell.terry@urology.ufl.edu,
    bianjiang@ufl.edu,
    xujie@ufl.edu
}

\usepackage{bibentry}

\begin{document}
\nocopyright
\maketitle

\begin{abstract}
Generative models can enhance discriminative classifiers by constructing complex feature spaces, thereby improving performance on intricate datasets. Conventional methods typically augment datasets with more detailed feature representations or increase dimensionality to make nonlinear data linearly separable. Utilizing a generative model solely for feature space processing falls short of unlocking its full potential within a classifier and typically lacks a solid theoretical foundation. We base our approach on a novel hypothesis: the probability information (logit) derived from a single model training can be used to generate the equivalent of multiple training sessions. Leveraging the central limit theorem, this synthesized probability information is anticipated to converge toward the true probability more accurately. To achieve this goal, we propose the \textbf{Bernoulli-Gaussian Decision Block (BGDB)}, a novel module inspired by the Central Limit Theorem and the concept that the mean of multiple Bernoulli trials approximates the probability of success in a single trial. Specifically, we utilize Improved Denoising Diffusion Probabilistic Models (IDDPM) to model the probability of Bernoulli Trials. Our approach shifts the focus from reconstructing features to reconstructing logits, transforming the logit from a single iteration into logits analogous to those from multiple experiments. We provide the theoretical foundations of our approach through mathematical analysis and validate its effectiveness through experimental evaluation using various datasets for multiple imaging tasks, including both classification and segmentation.
\end{abstract}

%
\section{Introduction}    
Classifiers are fundamental tools in machine learning, responsible for discerning intricate relationships between predictors and responses to allocate new observations into predetermined classes~\cite{rubinstein1997discriminative}. Among them, discriminative classifiers have gained prominence for their efficiency. Discriminative classifiers directly learn the conditional probability $P(y|x)$, selecting the label $y$ with the highest likelihood given an input $x$~\cite{raina2003classification,ng2001discriminative}.  This direct approach bypasses the need to model the joint probability distribution $P(x, y)$, as generative classifiers do, leading to faster decision-making~\cite{raina2003classification,ng2001discriminative}. Consequently, discriminative classifiers, particularly within convolutional neural networks (CNNs), have become the preferred choice for tasks such as image classification~\cite{krizhevsky2012imagenet}.


Despite their widespread use and efficiency, discriminative classifiers face challenges in extracting features and defining metric relations between examples, especially with complex data types such as medical images~\cite{jaakkola1998exploiting}. This limitation stems from their focus on learning the decision boundary rather than understanding the underlying data distribution. In contrast, generative models offer a promising solution by constructing more intricate feature spaces and providing a sophisticated framework for understanding the data generation process~\cite{perina2012free}. By creating structured hierarchies of latent variables linked through conditional distributions, generative models can establish nuanced correspondences between model components and observed features, enabling them to handle missing, unlabeled, and variable-length data effectively~\cite{perina2012free}. Techniques such as Fisher's method exemplify this approach, where original data is mapped into a low-dimensional feature space and then projected into a higher-dimensional space by kernel techniques for linear classification~\cite{jaakkola1998exploiting}. Another strategy involves augmenting data with generative models to improve feature representations, as seen in methods like Dataset Diffusion, which enhances the accuracy of segmentation and classification tasks~\cite{nguyen2024dataset}. However, the direct integration of generative models into feature construction in discriminative classifiers often lacks a robust theoretical foundation. In such cases, the generative model typically generates an unknown latent space from another unknown latent space, making the generation process inherently difficult to interpret.

In this paper, we propose a new hypothesis that the probability distribution obtained by a single training process can be used to generate the probability distribution for multiple training processes. Ideally, this generated distribution would represent the true classification probability distribution. Specifically, compared to other generative models such as GANs~\cite{goodfellow2014generative}, which produce data through the adversarial process between the generator and the discriminator, diffusion models~\cite{jarzynski1997equilibrium} have the advantage of generating one distribution from another and provide a mathematical foundation for this process. On the other hand, leveraging the distributions from a single training process, we can generate the probability distributions for multiple training iterations. According to the Central Limit Theorem, these generated distributions will more precisely approximate the true classification probabilities. This methodology thus enhances the model's classification performance through supervised learning. Building on this idea, we incorporated the diffusion model into the discriminative classifier, developing a Bernoulli-Gaussian Decision Block (BGDB) designed to enhance the deep learning model. Our contributions can be summarized as follows:


\begin{itemize}
    \item We introduce the Bernoulli-Gaussian Decision Block, which enhances the stability and performance of discriminative classifiers by leveraging the mean of logits from multiple experiments to supervise a single learning process.
    \item We employ IDDPM to construct and refine the probability distributions of Bernoulli Trials, improving inference accuracy without adding computational complexity during inference.
    \item We provide a theoretical analysis and validate the effectiveness of our approach through extensive experiments on multiple datasets, including Cityscapes, ISIC, and Pascal VOC, demonstrating notable improvements in classification and segmentation tasks.
\end{itemize}

\section{Related Work}
\subsection{Central Limit Theorem in Neural Networks}
Learning conditional and marginal probabilities from a dataset is fundamental to constructing machine learning methods, such as belief networks~\cite{davidson2004using}. Leveraging the central limit theorem (CLT) could enhance this process by providing a robust statistical foundation~\cite{davidson2004using}. According to the CLT, the sum of a large number of random variables approximates a Gaussian distribution. This principle also applies to neural networks, where the pre-activations of each layer tend to be Gaussian~\cite{lee2017deep,huang2021rethinking}. As the network width increases towards infinity, the output distribution of each neuron converges to a Gaussian distribution~\cite{zhang2022neural,lee2019wide}. Thus, optimization in neural networks can be framed as optimizing a Gaussian process~\cite{lee2017deep}.

Many neural network optimization techniques are developed based on the CLT. For instance, from a width-depth symmetry perspective, shortcut networks demonstrate that increasing the depth of a neural network also results in a Gaussian process manifestation~\cite{zhang2022neural}. In the Empirical Risk Minimization (ERM) framework, the long-term deviation, scaled by the CLT, is governed by a Monte Carlo resampling error, providing width-asymptotic guarantees independent of data dimension~\cite{chen2020dynamical}. Self-Normalizing Neural Networks utilize the CLT to approximate network inputs with a Gaussian distribution, enabling robust learning and introducing novel regularization schemes~\cite{klambauer2017self}. Despite these advancements, existing methods primarily rely on the CLT's mathematical properties for parameter estimation rather than directly modeling the CLT process within neural networks. This approach limits the potential of the CLT for optimizing neural networks to some extent.

\subsection{Logit-Based Optimization}
The logit function, introduced by Joseph Berkson in 1944, is derived from the term "logistic unit" and describes the logarithm of odds~\cite{berkson1944application, berkson1951prefer}. It maps the probability range $(0,1)$ to the entire real number line $(-\infty, +\infty)$, allowing the application of linear regression techniques to probabilities~\cite{cramer2003origins}. This mapping facilitates the use of regression methods in domains where outputs are naturally bounded probabilities rather than unbounded real numbers. In modern machine learning, the flexibility to let data drive model structures has led to more adaptive and predictive capabilities~\cite{zhao2020prediction}. This flexibility contrasts with traditional logit models, which often rely on specific data structures and inherent behavioral assumptions. 

Various methods have been developed to optimize neural networks by focusing on the logit function. Wu et al.~\cite{wu2021logit} introduced a reliable uncertainty measure based on logit outputs, aiding classification models in identifying instances prone to errors. This uncertainty measure can trigger expert intervention during high uncertainty classifications~\cite{wu2021logit}. Neural networks often exhibit overconfidence, producing high confidence scores for both in- and out-of-distribution inputs. Wei et al.~\cite{wei2022mitigating} addressed this issue with Logit Normalization (LogitNorm), modifying the cross-entropy loss to enforce a constant vector norm on the logits during training. In medical image analysis, Hu et al.~\cite{hu2021data} proposed logit space data augmentation, adaptively perturbing logit vectors to enhance classifier generalizability and mitigate overfitting from limited training data. These methods demonstrate that optimizing based on logit can significantly enhance neural network performance on finite datasets. 

\subsection{Diffusion Probabilistic Models}
\label{Probability Theory in Machine Learning}
Diffusion probabilistic models (DPMs) (or diffusion models [DMs]), inspired by non-equilibrium statistical physics~\cite{jarzynski1997equilibrium}, have recently gained traction in computer vision due to their remarkable generative capabilities. DMs generate highly detailed and diverse examples by iteratively reconfiguring data distribution through a diffusion process~\cite{croitoru2023diffusion, yang2023diffusion}. Incorporating small amounts of Gaussian noise, DMs use conditional Gaussians for straightforward parameterization of neural networks. Leveraging variational inference via a parameterized Markov chain~\cite{gagniuc2017markov}, DMs generate samples closely following the original data distribution within finite iterations.

Notable examples include latent diffusion models (LDMs)~\cite{rombach2022high, croitoru2023diffusion, yang2023diffusion}, which have set new standards in generative modeling. Stable Diffusion, a variant of LDMs, generates high-quality images based on text prompts, showcasing minimal artifacts and strong alignment with the prompts~\cite{rombach2022high, croitoru2023diffusion, yang2023diffusion}. DMs have been extensively applied in image generation~\cite{song2020score, nichol2021improved, ho2020denoising}, super-resolution~\cite{rombach2022high}, inpainting~\cite{batzolis2021conditional}, and image-to-image translation~\cite{choi2021ilvr}. Additionally, the latent representations learned by DMs have proven effective in discriminative tasks like image segmentation~\cite{baranchuk2021label}, classification~\cite{zimmermann2021score}, and anomaly detection~\cite{pinaya2022fast}. This versatility underscores the potential of diffusion models in a broad range of applications, connecting them to the field of representation learning, which includes designing novel neural architectures and developing advanced learning strategies~\cite{croitoru2023diffusion, yang2023diffusion}.

\section{Methods}
\label{headings}
In this paper, we propose the Bernoulli-Gaussian decision block, a novel module inspired by the CLT, which utilizes IDDPMs~\cite{nichol2021improved} to model the probability of Bernoulli trials. We will first review the formulation of IDDPMs, followed by a detailed description of the proposed Bernoulli-Gaussian Decision Block built upon the IDDPMs.

\subsection{Improved Denoising Diffusion Probabilistic Models}
\label{Bernoulli-Gaussian Decision Block}
Denoising Diffusion Probabilistic Models (DDPMs)~\cite{ho2020denoising} have demonstrated superior sample generation quality, often surpassing other generative models like GANs~\cite{goodfellow2014generative} and VQ-VAE~\cite{van2017neural}. Improved DDPMs (IDDPMs)~\cite{nichol2021improved} build on DDPMs by incorporating learned variances, allowing sampling in fewer steps with minimal quality loss.
In DDPMs, given data distribution $x_0 \sim q(x_0)$, a forward noising process $q$ generates latent variables $x_1$ through $x_T$ by adding Gaussian noise at each time $t$ with variance $\beta_t \in (0, 1)$, as follows~\cite{nichol2021improved}:
\begin{equation}
\label{DDPM}
\begin{aligned}
&q(x_1,...,x_T|x_0):= \prod_{t=1}^{T}q(x_t|x_{t-1}),\ \ \\
&\text{where}\ \ q(x_t|x_{t-1}):= \mathcal{N}(x_t;\sqrt{1-\beta_t} x_{t-1},\beta_t\boldsymbol{I}).
\end{aligned}
\end{equation}
With a sufficiently large $T$ and a carefully designed schedule for $\beta_t$, the latent variable $x_T$ approximates an almost isotropic Gaussian distribution~\cite{nichol2021improved}. Consequently, if the exact reverse distribution $q(x_{t-1}|x_t)$ were known, we could sample $x_T \sim \mathcal{N}(0,\boldsymbol{I})$ and reverse the process to obtain a sample from $q(x_0)$. However, since $q(x_{t-1}|x_t)$ relies on the entire data distribution, it is approximated using a neural network~\cite{nichol2021improved}:
\begin{equation}
\label{equ::2}
\begin{aligned}
&p_\theta(x_{t-1}|x_t) := \mathcal{N} (x_{t-1};\mu_\theta(x_t,t),\Sigma_\theta(x_t,t)), \ \ \\
&\text{where} \ \ \Sigma_\theta(x_t,t) = \sigma^2_{t}\textbf{I}.
\end{aligned}
\end{equation}

Through Maximum Likelihood Estimation (MLE), the distribution of $x_0$ can be derived. 
The combined use of $q$ and $p$ forms a variational auto-encoder, and the Variational Lower Bound (VLB) can be written as follows~\cite{nichol2021improved}:
\begin{equation}
\begin{aligned}
\label{equ::4}
L_{\text{vlb}} &= -\overbrace{\log p_\theta(x_0|x_1)}^{L_0}+\overbrace{D_{KL}(q(x_T|x_0)||p(x_T))}^{L_T}\\
&+ \sum_{t>1}\overbrace{D_{KL}(q(x_{t-1}|x_t,x_0)||p_\theta(x_{t-1}|x_t))}^{L_{t-1}}.
\end{aligned}
\end{equation}
With $\alpha_t:=1-\beta_t$ and $\bar{\alpha}_{t}:=\prod_{t}^{s=0}\alpha_s$, the marginal can be written as follow~\cite{nichol2021improved, ho2020denoising}:
\begin{equation}
\begin{aligned}
q(x_t|x_0) &= \mathcal{N}(x_t;\sqrt{\bar{\alpha}_t}x_0,(1-\bar{\alpha}_t)\boldsymbol{I}), \ \ 
\\
&\text{where}\ \ x_t=\sqrt{\bar{\alpha}_t}x_0+\sqrt{1-\bar{\alpha}_t}\epsilon, \ \epsilon \sim \mathcal{N}(0,\boldsymbol{I}).
\label{eq4}
\end{aligned}
\end{equation}

By applying Bayes' theorem, the posterior $q(x_{t-1}|x_t,x_0)$ can be determined with $\tilde{\beta}_t$ and $\tilde{\mu}_t(x_t,x_0)$, defined as follows~\cite{ho2020denoising,nichol2021improved}:
\begin{equation}
\begin{aligned}
&\tilde{\beta}_t:=\frac{1-\bar{\alpha}_{t-1}}{1-\bar{\alpha}_t}\beta_t, \ \ \ \ \\
&
\tilde{\mu}_t(x_t,x_0):=\frac{\sqrt{\bar{\alpha}_{t-1}}\beta_t}{1-\bar{\alpha}_t}x_0 + \frac{\sqrt{\alpha}_t(1-\bar{\alpha}_{t-1})}{1-\bar{\alpha}_t}x_t,\\
&q(x_{t-1}|x_t,x_0) = \mathcal{N}(x_{t-1};\tilde{\mu}_t(x_t,x_0),\tilde{\beta}_t\boldsymbol{I}).
\label{eq5}
\end{aligned}
\end{equation}


According to~\cite{ho2020denoising}, the $L_{t-1}$ can be calculated as:
\begin{equation}
\begin{aligned}
L_{t-1} &= \mathbb{E}_{ q(x_{1:T})}\left [\frac{1}{2\sigma_t^2}||\tilde{\mu_t}(x_t,x_0)-\mu_\theta(x_t,t)||^2\right ] + C.
\end{aligned}
\end{equation}
There are several ways to parameterize $\mu_\theta(x_t, t)$. One approach is to predict the noise $\epsilon$ with a neural network, and use Eqs.~(\ref{eq4}) and~(\ref{eq5}) to derive~\cite{ho2020denoising,nichol2021improved}:
\begin{equation}
\begin{aligned}
\mu_\theta(x_t,t) = \frac{1}{\sqrt{\alpha_t}}(x_t-\frac{\beta_t}{\sqrt{1-\bar{\alpha}_t}}\epsilon_\theta(x_t,t)).
\end{aligned}
\end{equation}
Predicting $\epsilon$ with a reweighted loss function has proven effective~\cite{ho2020denoising,nichol2021improved}:
\begin{equation}
\begin{aligned}
L_{\text{simple}} = \mathbb{E}_{t,x_0,\epsilon}[||\epsilon-\epsilon_\theta(x_t,t)||^2].
\end{aligned}
\end{equation}

In particular, as \cite{nichol2021improved} mentioned, IDDPM could generate a vector $v$ containing one component pre dimension, and this vector $v$ composes the new variances, $\Sigma_{\theta}(x_t, t)$ in Eq.~\ref{equ::2}:
\begin{equation}
\begin{aligned}
\Sigma_\theta(x_t,t) = \exp(v\log\beta_t+(1-v)\log \tilde{\beta}_t ).
\end{aligned}
\end{equation}
Since $L_{\text{simple}}$ doesn’t reply on $\Sigma_\theta(x_t,t)$~\cite{nichol2021improved}, the two loss functions $L_{\text{vlb}}$ and $L_{\text{simple}}$ can be simply combined into a new hybrid objective by introducing a hyperparameter $\lambda_{1}$ to scale one of them, i.e., 
\begin{equation}
L_{\text{hybrid}} = L_{\text{simple}} + \lambda_{1} L_{\text{vlb}}. 
\end{equation}
This reparameterization technique allows the diffusion model to reconstruct Gaussian distributions, enabling the transformation of the logit from a single iteration into logits analogous to those from
multiple experiments.

\subsection{Bernoulli Approximation}
\label{Bernoulli approximation}
In traditional settings, a single iteration of forward propagation yields one probability estimate. However, we can view each iteration as an independent and replicable trial, treating it as a Bernoulli Trial (BT). By conducting multiple independent trials within a single forward propagation, we can obtain more precise results. When the number of BTs is large enough, the distribution of the BT results approximates a Gaussian distribution, as described by the De Moivre–Laplace theorem~\cite{Walker2006DEMO}. This allows us to incorporate the CLT to estimate the mean of the Gaussian distribution, representing the results of BTs. This mean can be predicted, enabling us to simulate this Bernoulli process in a single iteration instead of multiple training runs.
 
A Bernoulli trial has exactly two possible outcomes: ``success'' (i.e., the positive case) and ``failure'' (i.e., the negative case). Let $p$ be the probability of the positive case. In a typical CNN, logits are generated and then converted into probabilities (for classification), confidence scores, and other expected outputs through functions like softmax and sigmoid. In an ideal scenario, the probability of the positive case $p=1$. Therefore, each training iteration can be viewed as a BT, with the logit representing the expected value of a random variable following the Bernoulli distribution. We define this random variable as the Bernoulli logit $y_{\text{Blogit}}$, which can take two fixed values: positive Bernoulli logit $y_{\text{Blogit}_+}$ and negative Bernoulli logit $y_{\text{Blogit}_-}$. The  logit $y_{\text{logit}}$ can be calculated using the following equation:
\begin{equation}
\begin{aligned}
y_{\text{logit}} &= \mathbb{E}(y_{\text{Blogit}}) = y_{\text{Blogit}_+} p + y_{\text{Blogit}_-} (1-p).
\end{aligned}
\end{equation}
If $p=1$, the logit is numerically equal to the positive Bernoulli logits, i.e., $y_{\text{logit}}=y_{\text{Blogit}_+}$. We refer to this process as the Bernoulli approximation.

Repeating the BT independently $n$ times, the possible values of the total number of positive outcomes range from $0$ to $n$. Let $\hat{p}$ denote the estimated probability of a positive outcome in $n$ trials, we have
\begin{equation}
\mathbb{E}(\hat{p}) = p, \hspace{0.8cm} Var(\hat{p}) = \frac{p(1-p)}{n}, 
\label{ylogit}
\end{equation}
where $\mathbb{E}(\hat{p})$ denotes the expected value of $\hat{p}$, $Var(\hat{p})$ denotes the variance of $\hat{p}$.
We incorporate a CNN to construct a Gaussian distribution by learning its mean and variance. According to the De Moivre–Laplace theorem~\cite{Walker2006DEMO}, as $n$ increases, the distribution of $\hat{p}$ increasingly resembles a Gaussian distribution:
\begin{equation}
\hat{p} \sim  \mathcal{N}(p,\sqrt{\frac{p(1-p)}{n} } ).
\label{mean_guassian}
\end{equation}

According to Eqs.~(\ref{mean_guassian}) and (\ref{ylogit}), the mean of the Gaussian distribution is numerically equal to the success probability of BT. Under optimal conditions, $y_{\text{logit}}$ can be calculated through multiple BTs. However, since the Bernoulli logit follows a Gaussian distribution, $y_{\text{logit}}$ can be calculated as follows:
\begin{equation}
\begin{aligned}
y_{\text{logit}} &= \mathbb{E}(y_{\text{Blogit}}) \\
&= y_{\text{Blogit}_+}\hat{p} + y_{\text{Blogit}_-} (1-\hat{p}).
\end{aligned}
\end{equation}
In an ideal scenario, the probability $\hat{p}$ is $1$, meaning each BT would succeed, otherwise is 0. Thus, the $y_{\text{logit}}$ is numerically equal to $y_{\text{Blogit}_+}$. Following Eq.~(\ref{mean_guassian}), after applying the softmax or sigmoid function, the mean of the Gaussian distribution can be used to categorize outputs as 0 or 1, thereby supervising the CNN model. Additionally, the variance of the Gaussian distribution would be zero in this ideal case, allowing us to simulate multiple BTs with their mean and variance in only one iteration. Through this entire process, logits are transformed into a Gaussian distribution.

\begin{figure*}[t]
\centering
  \includegraphics[width=0.85\textwidth]{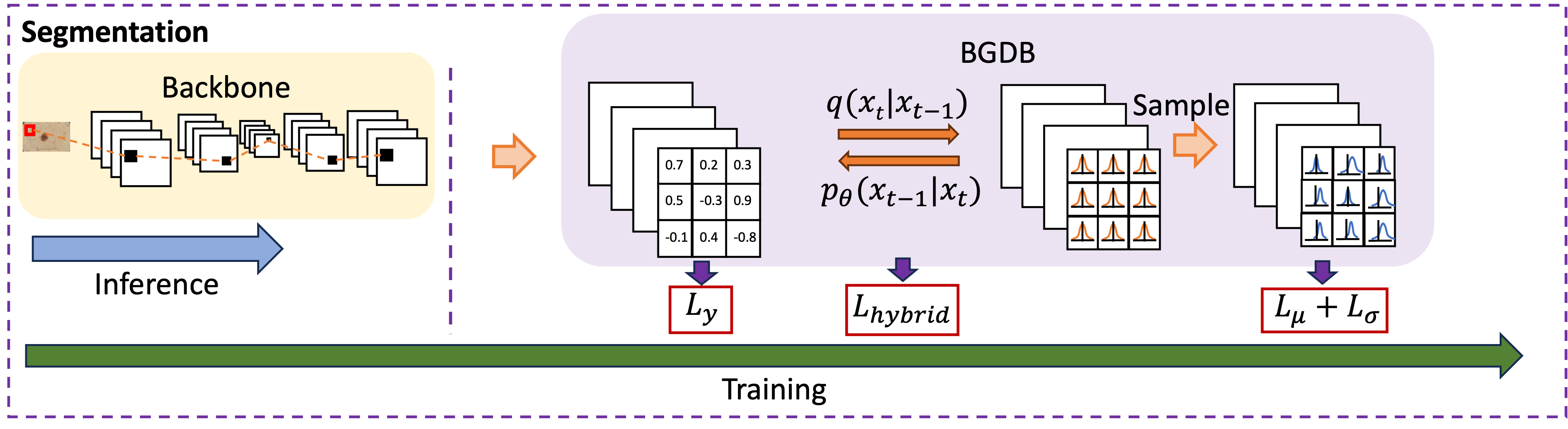}
  \caption{Workflow for performing segmentation tasks. The total loss in the training pipeline includes $L_y$ (task-specific loss), $L_{\text{hybrid}}$ (for IDDPM), and $L_\mu + L_\sigma$ (for Bernoulli approximation). After training, only the backbones are retained for inference.}
  \label{elbo_guassian}
\end{figure*}

\subsection{Bernoulli-Gaussian Decision Block}
Building on the concepts of Bernoulli approximation and IDDPMs, we introduce the Bernoulli-Gaussian decision block into the deep model training process, shown in Figure~\ref{elbo_guassian}. This Bernoulli-Gaussian Decision Block(BGDB) aims to enhance the stability and performance of discriminative classifiers by leveraging the mean of logits from multiple experiments to supervise a single learning process. Especially, $L_y$, task-specific loss such as Dice loss in segmentation tasks, can be calculated from the logit of a single learning process.

Meanwhile, we employ IDDPM to construct and refine the probability distributions of BTs. The entire construction process can be supervised by the $L_{\text{hybrid}}$. Compared to DDPM, IDDPM can generate both mean and variance, this approach perfectly aligns with Bernoulli Approximation. Simultaneously, through the inverse diffusion process, we sample the mean $\mu_{\text{output}}$ and variance $\sigma_{\text{output}}$ at time $t_0$, where $p(x_0) \sim (\mu_{\text{output}},\sigma_{\text{output}})$, from the logit produced by the backbone. After applying the softmax or sigmoid function, $\mu_{\text{output}}$ of the Gaussian distribution is required to categorize outputs as 0 or 1 to supervise the CNN model. Ideally, $\sigma_{\text{output}}$ should be $0$, allowing us to construct a multiple BTs with $\mu_{\text{output}}$ and $\sigma_{\text{output}}$ in a single iteration. 
Let $L_\mu$ and $L_\sigma$ be the loss targeting at mean $\mu_{\text{output}}$ and variance $\sigma_{\text{output}}$ (for Bernoulli approximation). Let $L_{\text{BCE}}$ denote the Balanced Cross-Entropy (BCE) loss, $L_{\text{MSE}}$ denote the Mean Squared Error (MSE) loss, $F$ represents the softmax or sigmoid function. Given that the mean is represented as a probability while the variance is numerically zero, the mean loss is calculated using BCE, whereas the variance loss is obtained using MSE. 

Thus, the entire loss function $\mathcal{L}$ for the model with BGDB module is calculated as follows:
\begin{equation}
\begin{aligned}
\mathcal{L} &= L_\text{y} + \lambda_{2} L_{\text{hybrid}} \\
+ \lambda_{3}&\left((\overbrace{ L_{\text{BCE}}F(\mu_{\text{output}}), \text{ label})}^{L_\mu} + \overbrace{L_{\text{MSE}}(\sigma_{\text{output}},0)}^{L_\sigma}\right). 
\end{aligned}
\label{loss}
\end{equation}

Since $F(\mu_{\text{output}})$ is a probability, it can also be used in other loss functions, such as Dice loss~\cite{milletari2016v}. This module is added after the logits and before the softmax to compute the loss function during training. After training, this structure is removed, and predictions are made using the original network, without any burden in inference. The label may encompass options such as the category of a single object or pixel.

This construction process begins by minimizing a loss function to generate a new distribution from an existing one. Because the output derived from the loss function adheres to the same distribution as the input, supervised learning is primarily needed for the mean and variance of the noise. By controlling these parameters, the entire diffusion process can transform one distribution into another desired distribution, the probability of multiple successful BT experiments. In generative tasks, the input distribution for diffusion models is initially fixed. However, in classification problems, the input logits are obtained through supervised learning, which can introduce instability. By leveraging the learning process of diffusion models, we use the distribution of logits from multiple experiments to supervise the logits obtained from a single training session. This approach aims to stabilize and enhance training by supervising the process with multiple experimental logits derived from a single training instance.

\section{Experiment}
\label{Experiments}
We evaluate the proposed method across various imaging tasks, including both classification and segmentation.

\subsection{Urban and General Scene Segmentation}
\paragraph{Datasets}
We utilized \textbf{Cityscapes}~\cite{cordts2016cityscapes} and PASCAL Visual Object Classes (VOC) Challenge (\textbf{Pascal VOC})~\cite{everingham2010pascal} datasets for this task. 
The Cityscapes dataset addresses the need for detailed semantic understanding by providing annotated stereo video sequences from 50 cities. It includes 5,000 images with high-quality pixel-level annotations, making it well-suited for evaluating segmentation methods that leverage extensive, high-quality labeled data. The Pascal VOC dataset offers publicly accessible images and annotations along with standardized evaluation software. 
For segmentation tasks, each test image requires predicting the object class for each pixel, with ``background'' designated for pixels that do not belong to any of the twenty specified classes. 


\paragraph{Compared Methods}
For our experiments on the Cityscapes and Pascal VOC datasets, we utilized the DeepLabV3 framework~\cite{chen2017rethinking,chen2018encoder}, following the experimental protocols outlined in the original papers. We evaluated the performance using three distinct backbones: MobileNet~\cite{howard2017mobilenets}, ResNet101~\cite{he2016deep}, and HRNet~\cite{wang2020deep}. This approach allowed us to systematically assess the model's adaptability and efficacy across varied scenarios.

\paragraph{Experimental Settings}
Our training regimen consisted of 30,000 iterations, with each batch comprising 16 samples. All input images were uniformly cropped to dimensions of $256 \times 256$. We employed the cross-entropy loss function, coupled with a learning rate of 0.01 and a weight decay of 1e-4. Stochastic Gradient Descent (SGD)~\cite{robbins1951stochastic} was used as the optimizer throughout the training process to ensure optimal convergence and model refinement. For testing, the images from the Cityscapes dataset retained their original size, while the Pascal VOC images were resized to $256 \times 256$.
Model performance was assessed using the Mean Intersection over Union (mIoU) metric.

In this study, all models were trained on an NVIDIA A100 GPU with 80 GB of memory. The hyperparameters were set as follows: $\lambda_1$ to $1 \times 10^{-3}$, and both $\lambda_{2}$ and $\lambda_{3}$ to 1. These settings were used for all subsequent experiments.

\paragraph{Experimental Results}
As illustrated in Table \ref{tab:natural_seg}, on both Cityscapes and Pascal VOC, all models experienced moderate improvements. Specifically, the models showed an increase in performance ranging from 0.08\% to 1.48\% on the Cityscapes dataset and from 0.21\% to 0.41\% on the Pascal VOC dataset. These results demonstrate the effectiveness of the proposed Bernoulli-Gaussian decision block in enhancing the performance.

\begin{table}[htbp]
  \centering
  \small
    \begin{tabular}{lll}
    \toprule
    \multicolumn{1}{c}{\multirow{2}[2]{*}{\textbf{Model}}} & \textbf{Cityscapes}& \textbf{Pascal VOC} \\ 
\cline{2-3}
& mIoU (\%) &  mIoU (\%) \\
    \hline
    DLP\_MobileNet 
    & 63.61 $\pm$ 0.72 & 61.78 $\pm$ 0.57\\
    \ \ \ \ +ours & 65.09 $\pm$ 0.38 \textcolor[rgb]{ 1,  0,  0}{+1.48} & 62.17 $\pm$ 0.65 \textcolor[rgb]{ 1,  0,  0}{+0.39} \\
    \hline
    DLP\_ResNet101
    & 72.00 $\pm$ 0.36 & 69.74 $\pm$ 0.49 \\
    \ \ \ \ +ours & 72.08 $\pm$ 0.10 \textcolor[rgb]{ 1,  0,  0}{+0.08}       & 69.95 $\pm$ 0.52 \textcolor[rgb]{ 1,  0,  0}{+0.21} \\
    \hline
    DLP\_HRNet 
    & 72.09 $\pm$ 0.47 & 69.87 $\pm$ 0.42\\
    \ \ \ \ +ours & 72.92 $\pm$ 0.37 \textcolor[rgb]{ 1,  0,  0}{+0.82} & 70.28 $\pm$ 0.57 \textcolor[rgb]{ 1,  0,  0}{+0.41} \\
    \bottomrule
    \end{tabular}%
    \caption{The results of mIoU (Mean $\pm$ Std) for urban and general scene segmentation on Cityscapes and Pascal VOC datasets. ``DLP'' denotes ``DeepLabv3+''.}
      \label{tab:natural_seg}
\end{table}%

\begin{table*}[htbp]
  \centering
  \small
    \begin{tabular}{cccccc}
    \toprule
    UNETR
    & FCN
    & U-Net
    & ResUNet
    & A*U-Net
    & \multicolumn{1}{c}{U-Net++
    } 
    \\
    77.62 $\pm$ 2.7 & 73.52 $\pm$ 2.7    & 69.17 $\pm$ 1.9 & 75.82 $\pm$ 1.24 & 72.47 $\pm$ 1.86 & \multicolumn{1}{c}{80.78 $\pm$ 0.83 } \\
    \hline
    \ \ \ \ +ours & \ \ \ \ +ours  & \ \ \ \ +ours & \ \ \ \ +ours & \ \ \ \ +ours & \multicolumn{1}{c}{\ \ \ \ +ours} \\
    80.30 $\pm$ 2.45 & 75.04 $\pm$ 2.7    & 73.91 $\pm$ 1.19 & 76.44 $\pm$ 0.84 & 73.49 $\pm$ 0.98 & \multicolumn{1}{c}{80.49 $\pm$ 1.22 } \\
    \hline
    \textcolor[rgb]{ 1,  0,  0}{+2.68} & \textcolor[rgb]{ 1,  0,  0}{+1.52} & \textcolor[rgb]{ 1,  0,  0}{+4.74} & \textcolor[rgb]{ 1,  0,  0}{+0.62} & \textcolor[rgb]{ 1,  0,  0}{+1.02} & \textcolor[rgb]{ 0,  0,  1}{-0.29} \\
    \bottomrule
    \end{tabular}%
    \caption{Dice (Mean (\%) $\pm$ Std) for skin lession segmentation on ISIC dataset. ``A*U-Net'' denotes ``Attention U-Net''.}
    \label{tab:ISIC}%
\end{table*}%

\subsection{Skin Lesion Segmentation}
\paragraph{Datasets} We used the International Skin Imaging Collaboration (ISIC) dataset~\cite{tschandl2018ham10000,codella2019skin} for skin lesion segmentation. The ISIC dataset is the world's largest collection of dermoscopic skin images. The ISIC 2018 challenge, held at the MICCAI conference, included three tasks and featured over 12,500 images. The challenge attracted 900 registered users, with 115 submissions for lesion segmentation, 25 for lesion attribute detection, and 159 for disease classification. 

\paragraph{Compared Methods} 
We evaluated the Bernoulli-Gaussian decision block across several classical and state-of-the-art 2D medical segmentation models using the ISIC dataset. These models include U-Net~\cite{ronneberger2015u}, Attention U-Net~\cite{oktay2018attention}, U-Net++~\cite{zhou2019unet}, FCN~\cite{liu20183d}, ResUNet~\cite{diakogiannis2020resunet}, and UNETR~\cite{hatamizadeh2022unetr}, all implemented using the MONAI framework~\cite{cardoso2022monai}. The baseline models were trained using the Dice loss~\cite{milletari2016v}, while ``+ours'' models were trained with our proposed loss function in addition to the Dice loss.

\paragraph{Experimental Settings}
We utilized the training, validation, and test datasets provided by the ISIC 2018 challenge. These datasets were combined and then randomly split into training and testing sets in a 5:2 ratio (2,600 images for training and 1,094 for testing). We performed 5-fold cross-validation, selecting the optimal model from each fold's validation set. The selected models were then evaluated on the testing set, and we recorded the mean and variance of performance metrics across the 5 folds. For data augmentation, we normalized pixel values to a range between 0 and 255 and resized the images to $256 \times 256$ to meet the input requirements of the proposed block. 

The models were trained using the AdamW~\cite{loshchilov2017decoupled} optimizer with a weight decay of $1e-5$ and a learning rate of $1e-4$. Each model underwent 10,000 iterations of training, with the goal of achieving the highest Dice scores. This approach enabled a thorough comparative analysis between the baseline and enhanced models by the proposed decision block. Model performance was assessed using Dice score~\cite{milletari2016v}, which measures the overlap between the predicted segmentation and the ground truth.

\paragraph{Experimental Results}
Table~\ref{tab:ISIC} shows the Dice scores for the models on the ISIC dataset. Upon integrating the proposed block, we observed performance improvements across most models, with the exception of U-Net++, which experienced a marginal decline of -0.29\%. The performance improvements for the other models ranged from 0.6\% to 4.74\%.


\subsection{Beyond Segmentation: Skin Lesion Classification}
\paragraph{Datasets and Compare Methods} Similar to the skin lesion segmentation task, we used the ISIC 2018 challenge dataset for skin lesion classification~\cite{tschandl2018ham10000,codella2019skin}. 
We conducted empirical analyses across a spectrum of prominent models to assess the efficacy of the Bernoulli-Gaussian decision block. The models included DenseNet~\cite{huang2017densely}, ResNet~\cite{he2016deep}, Vision Transformer (ViT)~\cite{dosovitskiy2020image}, EfficientNet~\cite{tan2019efficientnet}, and SENet~\cite{hu2018squeeze}.

\paragraph{Experimental Settings}
We utilized the entirety of the ISIC 2018 dataset, amalgamating all available images before randomly partitioning them into training and testing sets in a 5:1 ratio while maintaining the original distribution. We conducted rigorous 5-fold cross-validation within the test set. From each fold, we selected the model with the highest accuracy on the validation set for final testing. We documented the mean and variance of accuracy and the Area Under the ROC Curve (AUC) across the 5-fold models. 

To rigorously evaluate the model's performance, we employed basic data augmentation strategies, including random rotations up to 15 degrees, flipping, and zooming in or out by a scale of 0.1 with a 50\% probability. We used the Adam optimizer~\cite{kingma2014adam} with a learning rate of 1e-5. Each model was trained for 50 epochs to achieve the highest levels of accuracy. This approach allowed for an exhaustive comparative analysis between models with and without the proposed block, enhancing our understanding of their respective performances. 
\begin{table}[htbp]
  \centering
  \small
    \begin{tabular}{lll}
    \toprule
    \multicolumn{1}{c}{\multirow{2}[2]{*}{\textbf{Model}}}&\multicolumn{2}{c}{\textbf{ISIC}} \\
    \cmidrule(r){2-3}
          & Accuracy (\%) & AUC (\%)  \\
    \hline
    DenseNet169
    & 68.34 $\pm$ 0.59 & 89.36 $\pm$ 0.50 \\
    \ \ \ \ +ours & 69.83 $\pm$ 1.05 \ \textcolor[rgb]{ 1,  0,  0}{+1.49}      & 90.10 $\pm$ 0.49\ \textcolor[rgb]{ 1,  0,  0}{+0.74} \\
    \hline
    ViT
    & 69.28 $\pm$ 0.59 & 89.06 $\pm$ 0.24  \\
    \ \ \ \ +ours & 69.24 $\pm$ 0.65 \ \textcolor[rgb]{ 0,  0,  1}{-0.04}    & 89.30 $\pm$ 0.18\ \textcolor[rgb]{ 1,  0,  0}{+0.24}  \\
    \hline
    ResNet50
    & 66.66 $\pm$ 0.90 & 88.44 $\pm$ 0.29\\
    \ \ \ \ +ours & 68.26 $\pm$ 1.01 \ \textcolor[rgb]{ 1,  0,  0}{+1.60}    & 88.90 $\pm$ 0.85\ \textcolor[rgb]{ 1,  0,  0}{+0.46}  \\
    \hline
    SENet154
    & 69.64 $\pm$ 1.04  & 89.79 $\pm$ 0.38  \\
    \ \ \ \ +ours & 70.18 $\pm$ 1.50 \ \textcolor[rgb]{ 1,  0,  0}{+0.54}    & 90.23 $\pm$ 0.56\  \textcolor[rgb]{ 1,  0,  0}{+0.44} \\
    \hline
    EfficientNet
    & 65.78 $\pm$ 0.68 & 88.44 $\pm$ 0.49 \\
    \ \ \ \ +ours & 67.14 $\pm$ 0.60\ 
 \textcolor[rgb]{ 1,  0,  0}{+1.36} & 89.10 $\pm$ 0.21\ \textcolor[rgb]{ 1,  0,  0}{+0.66}  \\
    \bottomrule
    \end{tabular}%
      \caption{Accuracy and AUC (Mean (\%) $\pm$ Std) for skin lession classification on ISIC dataset.}
        \label{tab:ISIC_classification}
\end{table}%

\paragraph{Experimental Results}
Table~\ref{tab:ISIC_classification} shows the accuracy and AUC scores for the models on the ISIC dataset. The experimental findings indicate that, aside from slight declines in the ViT (0.04\% accuracy), all other models exhibited performance enhancements. The improvements ranged from 0.54\% to 1.6\% in accuracy and from 0.24\% to 0.74\% in AUC.

\subsection{Ablation Experiments}
To thoroughly understand the impact of different loss functions on the performance of our model, we conducted ablation experiments using the U-Net model with various combinations of loss functions. We set all hyperparameters to 1. The loss functions evaluated included the task-specific loss (i.e., Dice loss $L_{\text{Dice}}$), the diffusion loss $L_{\text{hybrid}}$ for IDDPM, the BT loss for Bernoulli approximation (i.e., $L_\mu + L_\sigma$). The specific combinations tested were: U-Net (i.e., Only $L_{\text{Dice}}$), $L_{\text{Dice}}+L_\mu+L_\sigma$ (i.e., no diffusion loss), $L_{\text{Dice}}+ L_{\text{hybrid}}$ (i.e., no BT loss),  all losses combined (i.e., $L_{\text{Dice}}+L_\mu+L_\sigma+L_{\text{hybrid}}$). For each combination of loss functions, we trained the U-Net model on the ISIC dataset using the same experimental settings as described previously. 

Table~\ref{tab:ablation} shows the Dice scores for the different combinations of loss functions. Our analysis revealed that incorporating all loss functions led to the best performance, with a Dice score improved by 4.74\%. This underscores the profound impact of combining multiple loss functions on model performance. 

\begin{table}[htbp]
  \centering
  \small
    \begin{tabular}{cccc}
    \toprule
    \multicolumn{1}{l}{U-Net} & \multicolumn{1}{l}{No diffusion loss} & \multicolumn{1}{l}{No BT loss} & \multicolumn{1}{l}{All losses combined} \\
    \midrule
    69.17 & 72.23 & 73.17 & \textcolor[rgb]{ 1,  0,  0}{73.91} \\
    \bottomrule
    \end{tabular}%
    \caption{Dice scores of different loss combinations for skin lession segmentation on the ISIC dataset.}
    \label{tab:ablation}%
\end{table}%

\begin{figure}[t]
\centering
  \includegraphics[width=0.4\textwidth]{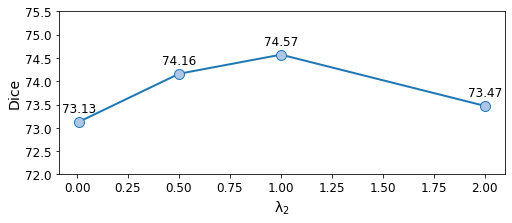}
  \includegraphics[width=0.4\textwidth]{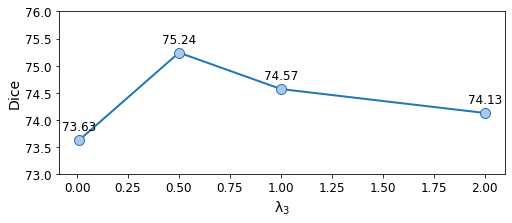}
  \caption{Impact of hyperparameters $\lambda_2$ and $\lambda_3$ on segmentation model performance, with the other hyperparameters fixed at 1.
  }
  \label{coefficient}
\end{figure}

We also explored the impact of hyperparameter settings of $\lambda_2$ and $\lambda_3$ in Eq.~\ref{loss}. This ablation study was conducted on the first fold of the U-Net experiment's dataset. The hyperparameter preceding the initial model's Dice loss was fixed at 1. Initially, with $\lambda_3$ set to 1, the hyperparameter before the $\lambda_2$ was varied at 0.01, 0.5, 1, and 2 to observe changes in model performance. Similarly, with the Dice loss and $\lambda_3$ fixed at 1, the $\lambda_2$ was adjusted to 0.01, 0.5, 1, and 2, allowing us to evaluate its effect on the model's performance. Figure~\ref{coefficient} shows that the model performs most stably when the hyperparameters for all three losses are consistent, which aligns with our default setting. Additionally, fine-tuning the hyperparameters within the range of 0.5 to 1 can be beneficial for achieving optimal performance.

\section{Discussion}
While our model achieved moderate advancements in segmentation tasks, it encounters several limitations. Statistically, the proposed Bernoulli-Gaussian decision block relies on a sufficiently large number of trials, $n$, to satisfy the formula under optimal conditions. The block can determine $n$ to ensure the validity of the mean of the Gaussian distribution.

The IDDPM, despite its faster training and inference speeds compared to DDPM, experiences a slowdown due to the simultaneous training and inference processes within the proposed block. This limitation restricts the time step of the diffusion model to a minimum of 25, posing challenges in model training. Additionally, our experiments focused on 2D segmentation with image dimensions of $256 \times 256$. Performance optimization could involve adjusting input image dimensions and model depth, but the deceleration in training speed has hindered further refinement. The extensive parameter count of IDDPM also makes it impractical for handling 3D images, necessitating very small image sizes, which is unsuitable for 3D segmentation tasks.

The U-Net architecture's encoder-decoder structure limits the predicted value dimensions to powers of 2, complicating classification tasks. On the ISIC dataset, we explored resizing logits from $1 \times 1$ to $64 \times 64$ and averaging the reconstructed results from IDDPM. Although the model shows potential for classification, the averaging operation appears redundant. Altering the U-Net to better handle prediction noise could improve its application to classification tasks. Moreover, considering the complexity of IDDPM, our BGDB block uses default hyperparameters that are suitable for the generative model. To fully harness the potential of this model, a meticulous tuning process is required. This process needs to be based on more simple BGDB block to be effectively conducted.

\section{Conclusion}
We proposed a novel Bernoulli-Gaussian Decision Block, which involves constructing multiple experimental probability distributions using the diffusion model. This method achieved modest performance improvements in segmentation tasks and showed promising potential for addressing classification challenges. 

\bibliography{aaai25}
\end{document}